# EmoScan: Automatic Screening of Depression Symptoms in Romanized Sinhala Tweets

Jayathi Hewapathirana
Informatics Institute of Technology
Colombo, Sri Lanka
jayathi.2019278@iit.ac.lk

Deshan Sumanathilaka
Informatics Institute of Technology
Colombo, Sri Lanka
deshan.s@iit.ac.lk

*Abstract*— This work explores the utilization of Romanized Sinhala social media data to identify individuals at risk of depression. A machine learning-based framework is presented for the automatic screening of depression symptoms by analyzing language patterns, sentiment, and behavioural cues within a comprehensive dataset of social media posts. The research has been carried out to compare the suitability of Neural Networks over the classical machine learning techniques. The proposed Neural Network with an attention layer which is capable of handling long sequence data attains a remarkable accuracy of 93.25% in detecting depression symptoms, surpassing current state-of-the-art methods. These findings underscore the efficacy of this approach in pinpointing individuals in need of proactive interventions and support. Mental health professionals, policymakers, and social media companies can gain valuable insights through the proposed model. Leveraging natural language processing techniques and machine learning algorithms, this work offers a promising pathway for mental health screening in the digital era. By harnessing the potential of social media data, the framework introduces a proactive method for recognizing and assisting individuals at risk of depression. In conclusion, this research contributes to the advancement of proactive interventions and support systems for mental health, thereby influencing both research and practical applications in the field.

## I. INTRODUCTION

Depression, a prevalent and serious mental health disorder with potentially severe consequences, including suicide, is emphasized by the American Psychiatric Association [1]. It is acknowledged as a significant medical disorder that impacts a person's feelings, actions, and thoughts and is linked to physical and emotional problems. Teenagers and the elderly, particularly those with late-onset depression, are at a high risk of suicide. According to the World Health Organization [2], suicide causes 800,000 deaths annually and is the second leading cause of death among individuals aged 15-29. The risk of depression has been further exacerbated by the COVID-19 pandemic, with its disruptions and challenges [3]. Recovering from the pandemic is particularly challenging for those affected, as it introduces new and unfamiliar lifestyles. Factors such as the stigma surrounding sadness, shyness, and embarrassment discourage people from seeking help for their mental health issues. Social media comprises interactive tools that enable users to create, share, and express information, ideas, and interests within online groups and networks [4].

Individuals utilize social media to connect with friends, family, communities, while businesses employ it to monitor customer feedback, market their products, and gather customer intelligence [5]. Social media, accessible through mobile applications, enjoys immense popularity worldwide [6]. Well-known social networking sites such as Twitter, Facebook, Reddit, and LinkedIn are characterized by their popularity. Users tend to exhibit a greater inclination to express their thoughts, engage with others, and share their emotions in their native language [7]. In Sri Lanka, where Sinhala is widely spoken, the introduction of Romanized Sinhala support on social media platforms has resulted in an encouragement for users to express themselves in their native language [8]. Therefore, a significant volume of social data is generated, which reveals insights into people's interests, moods, and behaviour.

Sinhala, despite being a morphologically rich but under-resourced language for natural language processing (NLP) in the field of artificial intelligence, has witnessed a significant increase in the availability of Sinhala text-based web content with the adoption of the Unicode character set [9]. This growth is reflected in the content posted on social media platforms, as social media becomes more popular in Sri Lanka [10]. Users on social media often resort to using the English script to write Sinhala words, known as Romanized Sinhala [9]. Consequently, a blend of Sinhala and English, including the utilization of "Singlish" (Sinhala written in the Roman script), can be observed [10]. Analyzing sentiment based solely on the Sinhala language may result in erroneous interpretations, as the presence of English or Singlish content can influence the sentiment of the data [9]. Challenges arise with traditional methods of detecting mental health issues, particularly in terms of diagnosis and tracking. To tackle this, online screening tools that leverage machine learning techniques have proven to be valuable in identifying symptoms of mental health problems, and their utilization is anticipated to increase in future evaluation approaches. The contribution of this can be depicted as follows.

- Presents a novel Neural Network architecture with an attention layer, achieving a remarkable 93.25% accuracy in detecting depression symptoms within Romanized Sinhala social media data.
- Demonstrates the potential of social media analytics, NLP, and machine learning for proactive mental health screening, offering valuable insights for practitioners in developing early intervention and support systems.

Section II of our study presents an overview of existing literature in our research domain and discusses the limitations of previous works. We highlight the importance and necessity of our work in addressing these limitations. The Methodology and Materials section provides a detailed explanation of the dataset utilized and the workflow followed in our research.

We describe the specific methods and techniques employed in our study. Moving on to the Results section, the findings and outcomes of our research are presented and illustrated. The effectiveness and performance of our approach are demonstrated through these results. Finally, in Section V, the concluding remarks of our study are presented. The key findings, contributions, and implications of our research are summarized, emphasizing its significance in the broader context of the research domain.

## II. RELATED WORK

In a recent study conducted by [11], an automated multimodal framework was introduced for anticipating potential depressed consumers using self-report comments. The framework utilized a dataset consisting of Tweets. A notable advancement in their approach was the selection of features based on correlation. Nine different classification models, including SVM, LR, DT, Gradient Boosting, RF, Ridge Classifier, AdaBoost, Catboost, and Multilayer Perceptron, were employed to analyze the features. However, it is important to acknowledge certain limitations of their study. The detection of depression in Romanized Sinhala or Sinhala posts is not possible with the framework, thus limiting its applicability to specific languages.

In a study conducted by [12], Reddit users' posts were utilized as a dataset to identify patterns that could indicate relevant online users' attitudes toward depression. The study made advancements in several areas. Firstly, three LIWC (Linguistic Inquiry and Word Count) features were designed to extract meaningful linguistic characteristics from the posts. Additionally, the effectiveness of using Ngrams probabilities, LIWC, and LDA (Latent Dirichlet Allocation) as individual features was assessed to accurately predict performance. The study focused on analyzing textual posts as the primary source of features. Various machine learning models, including LR, SVM, AdaBoost, RF, and Multilayer Perceptron, were employed for the classification task. Performance evaluation was done using metrics such as precision, recall, F1-score, and accuracy. In a study conducted by [13], the objective was to analyze depression using Facebook data obtained from a public internet source. Advancements were made by training a model to utilize three types of factors—emotional process, temporal process, and linguistic style—both independently and in combination. Different machine learning techniques, including DT, KNN, SVM, and Ensemble, were employed to investigate the effectiveness of depression detection. Performance evaluation was conducted using metrics such as precision, recall, and F1-score.

In a study conducted by [14], the focus was on examining the relationship between the content of uploaded Instagram pictures and the personality traits of users. To collect data, an online survey was conducted, wherein participants were asked to grant access to their Instagram account through the Instagram API. The study made advancements by analyzing the content of the uploaded pictures using the Google Vision API. Additionally, K-means clustering and Spearman's correlation analysis approaches were employed to explore the relationships between picture content and personality traits.

## III. METHODOLOGY

In this research, the focus is on the detection of signs of depression in Romanized Sinhala tweets and the distinction from ordinary content. The approach involves the creation of machine learning models using a dataset collected from sources like Twitter. The dataset is manually labeled as indicating depression or not, and it is utilized for both the training and testing phases. The origin and annotation process of the dataset are described, followed by detailed steps for data preprocessing and feature extraction. The experimental setup design includes the selection of appropriate algorithms for supervised classification, such as Naive Bayes, SVM, Random Forest, or Deep Learning models like RNNs. By following this methodology, an effective system for the identification of depression in Romanized Sinhala tweets is aimed to be developed, contributing to the provision of mental health support in the Sinhala-speaking community.

### A. Data Set Collection

A successful experiment on depression detection requires the presence of a labelled corpus, which is of crucial importance. The dataset collection process involved the gathering of English depression and non-depression tweet data from sources such as Kaggle and articles. The data was subsequently translated into Sinhala using the Google Translate API. Following that, the Sinhala dataset was converted into Romanized Sinhala using a rule-based transliteration approach [15] and Swa-Bhasha Transliterator [16].

Through this multistep process, a comprehensive dataset comprising tweets related to depression and non-depression in Romanized Sinhala languages was created.

TABLE I. DATASET ARRANGEMENT

| Tweets | Tag | Nature |
|---|---|---|
| mata thiyena lokuma prashnaya hemade genama hithana eka. | 1 | depressive |
| mata oyata therum karanna be mage ethule wena dewal katawath therum karanna denna be mata eka pehedili karannawath be | 1 | depressive |
| oya hondin inna mama ne eth kamak ne | 1 | depressive |
| mama ada ude jim ekata gihin yoga kala. | 0 | non-depressive |

### B. Manually Annotations of Dataset

To detect depression in Romanized Sinhala tweets, a dataset consisting of two tags, "depression" and "non-depression," was compiled. A CSV file was created to store this data, with two columns, "text" and "label." The "label" column was assigned a label of "1" for depressive tweets and "0" for non-depressive tweets. The dataset was manually annotated during the annotation process. It comprised a total of 6014 entries, with 2997 annotations for depression tweets and 3017 annotations for non-depression tweets. The dataset can be accessed through Kaggle [17].

### C. Preprocessing

Text preprocessing plays a crucial role in preparing the Romanized Sinhala tweets for analysis. Firstly, the removal of non-Sinhala characters is performed to ensure the focus is solely on the Sinhala content, with punctuation marks and symbols being eliminated. Next, stop words, which lack significant meaning despite their common usage, are removed to reduce noise in the data. Lastly, stemming is applied to reduce the dimensionality of the feature vector by transforming words into their base or root form. These preprocessing techniques contribute to data streamlining, improved analysis efficiency, and enhanced accuracy of subsequent classification models for the detection of depression.

*D. Feature Extraction*

The Python toolkit, sklearn, is utilized for feature extraction. Features are extracted from the text data using modules such as CountVectorizer and TfidfVectorizer. Different n-gram feature types, including unigrams, bigrams, and trigrams, are explored to capture diverse contextual information. Feature selection is performed using the SelectKBest function, which applies the chi-square score. The labels are encoded using LabelEncoder. The pipeline, consisting of feature extraction and feature selection, is fitted on the training data, and the resulting selected features are transformed into an array. The purpose of this code is to extract informative features from the text data and prepare it for classification. By employing various feature types and selection techniques, the goal is to enhance the accuracy, precision, recall, and F1 measure of the classification model. The code also computes the confusion matrix and generates a classification report to evaluate the performance of the model.

*E. Classifications*

Depression detection in Romanized Sinhala tweets was performed using supervised learning techniques. The performance of four classification algorithms, including Neural Network, SVM, Decision Trees, Random Forest Classifier, and Gaussian Naive Bayes classifier, was compared. The aim was to train models that could classify tweets as either indicative of depression or not. The best-performing Neural model architecture is defined using the Keras library by creating a Sequential model and adding two dense layers to it. The first dense layer, which uses the 'relu' activation function, has 512 nodes and an input dimension of the number of columns in the 'selected_features' dataset. A Dropout layer is also added after the first dense layer to prevent overfitting. Finally, a dense layer with a single node and 'sigmoid' activation function is added to produce binary classification output. The model is compiled by using the 'binary_crossentropy' loss function, 'adam' optimizer, and 'accuracy' metric.Evaluation metrics such as accuracy, precision, recall, and F1 score were utilized for the quantitative evaluation of the models. The objective was to determine the algorithm(s) with the highest accuracy in detecting depression in Romanized Sinhala tweets.

*F. Evaluation*

The data is split into training and test sets, and the model is trained on the training set. Probabilities are predicted for the test set and converted into class predictions using a threshold of 0.5. The evaluation metrics are calculated based on the predicted classes and true labels of the test set.

Fig 1 shows the data pipeline used in the final prototype to identify the depression status based on the user's chat.

## IV. RESULTS AND FINDINGS

In the results and evaluation section, a comparison was made between four different classifiers and different feature sets based on their accuracy, precision, recall, and F-score measures. The tweets were stored in a single CSV file, and both the training and testing data were also stored in a separate

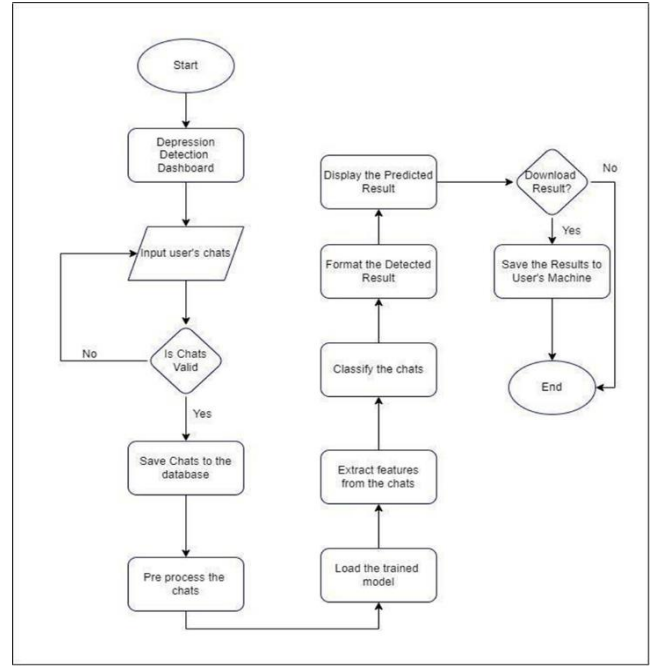

Fig. 1. Data flow in the proposed Model

CSV file. To divide the data into training and testing sets, a function was used, assigning 80% of the data to the training set and 20% to the test set. Features were extracted using CountVectorizer and Tf-idf Vectorizer from the sklearn toolkit. The same feature vector was then used as input for each of the four different algorithms. The performance of the models was evaluated using various metrics including F1 score, precision, recall, and accuracy, utilizing the testing data.

*A. Evaluation Matrix*

For rigorous evaluation of the proposed model, we employed several performance metrics. Accuracy, the proportion of correct predictions, was the fundamental measure. To address potential class imbalances, the F1-score was calculated. This metric offers a harmonic mean of precision (the fraction of true positives among predicted positives) and recall (the fraction of true positives correctly identified). Precision highlights the model's ability to minimize false positives, while recall indicates its capacity to find true positives. These metrics were computed using standard formulas and scikit-learn's functions for streamlined analysis.

Table 2 represents the Precision, Recall, F1-Score and support data for the train model. A benchmarking analysis was performed on various classification models, including the Decision Support Vector Machines (SVM), Tree Classifier, Random Forest Classifier, Gaussian Naive Bayes Classifier, and a Neural Network model. Logistic Regression, a linear classification model that estimates outcome probabilities using the logistic/sigmoid function, was also incorporated into the analysis. The Decision Tree Classifier constructed a decision tree model to determine outcomes, while the Random Forest Classifier utilized ensemble learning with multiple decision trees. The Gaussian Naive Bayes Classifier assumed independence among features and calculated probabilities. Upon evaluating the performance of each model, it was observed that the Neural Network model exhibited the highest accuracy score of 0.933, surpassing the other classifiers. The confusion matrix further demonstrated that the Neural Network model had the fewest false positives and false negatives, indicating its superior ability to accurately identify positive and negative instances. It is noteworthy that the Support Vector Machines (SVM) model achieved the second-highest accuracy score of 0.923, while the Decision Tree Classifier had the lowest accuracy score of 0.86 among the tested models. The Random Forest Classifier and Gaussian Naive Bayes Classifier achieved accuracy scores of 0.912 and 0.900, respectively. In conclusion, based on the evaluation of accuracy scores and the confusion matrix, the Neural Network model emerged as the top-performing model among the tested classifiers. It was selected as the preferred model for this specific classification problem due to its exceptional performance in accurately identifying positive and negative instances.

TABLE II. ACCURACY OF EACH STUDY

| Model | Accuracy | Label | Precision | Recall | F1-Score | Support FN/ FP |
|---|---|---|---|---|---|---|
| Neural Network | 0.93 | 0 | 0.94542254 | 0.91482112 | 0.92987013 | 540 |
| | | 1 | 0.92125984 | 0.94967532 | 0.9352518 | 582 |
| Support Vector Machines (SVM) | 0.92 | 0 | 0.93771626 | 0.92333901 | 0.9304721 | 542 |
| | | 1 | 0.928 | 0.94155844 | 0.93473006 | 580 |
| Decision Trees | 0.86 | 0 | 0.87024221 | 0.85689949 | 0.86351931 | 503 |
| | | 1 | 0.8656 | 0.87824675 | 0.87187752 | 541 |
| Random Forest Classifier | 0.91 | 0 | 0.90831919 | 0.91141397 | 0.90986395 | 535 |
| | | 1 | 0.91530945 | 0.91233766 | 0.91382114 | 562 |
| Gaussian Naive Bayes Classifier | 0.90 | 0 | 0.95694716 | 0.8330494 | 0.89071038 | 535 |
| | | 1 | 0.8583815 | 0.96428571 | 0.90825688 | 562 |

## V. CONCLUTION AND FUTURE WORKS

In this research, the detection of depression in Romanized Sinhala tweets using machine learning techniques was explored. Strong accuracy in identifying depressive tweets was demonstrated, particularly by the Neural Network model. Several future research directions are suggested to advance this field of study. Firstly, the analysis could be extended to include other social media platforms, enabling the connection of user accounts for personalized predictions and interventions based on their latest comments. Finally, the integration of the developed model into existing mental health resources is proposed to facilitate timely support and tailored recommendations for early intervention. By leveraging social media data and continuing research, mental health monitoring in the digital age can be effectively conducted, providing valuable assistance to individuals in need.